\documentclass[lettersize,journal]{IEEEtran}
\usepackage{amsmath,amsfonts}
\usepackage{algorithmic}
\usepackage{algorithm}
\usepackage{array}
\usepackage[caption=false,font=normalsize,labelfont=sf,textfont=sf]{subfig}
\usepackage{textcomp}
\usepackage{stfloats}
\usepackage{url}
\usepackage{verbatim}
\usepackage{graphicx}
\usepackage{subcaption}
\usepackage{cite}
\usepackage{booktabs}
\usepackage{multirow}
\usepackage{tabularx}
\usepackage{color}
\newcolumntype{C}{>{\centering\arraybackslash}X} 
\newcommand{\eg}{e.g., }
\newcommand{\ie}{i.e., }
\newcommand{\dataset}{\textit{DUTS-MM}}
\newcommand{\datasetq}{\textit{DUTS-MQ}}

\begin{document}
\title{Pluralistic Salient Object Detection}

\author{Xuelu~Feng, Yunsheng~Li, Dongdong~Chen, Chunming~Qiao,~\IEEEmembership{Fellow,~IEEE,} Junsong~Yuan,~\IEEEmembership{Fellow,~IEEE,} Lu Yuan, and Gang~Hua,~\IEEEmembership{Fellow,~IEEE}
\thanks{Xuelu Feng, Chunming Qiao, Junsong Yuan are with the Department of Computer Science and Engineering, University at Buffalo, USA (e-mail: xuelufen@buffalo.edu; qiao@buffalo.edu; jsyuan@buffalo.edu).}
\thanks{Yunsheng Li, Dongdong Chen, Lu Yuan are with Microsoft GenAI, USA (e-mail: yunshengli@microsoft.com; dongdong.chen@microsoft.com; luyuan@microsoft.com)}
\thanks{Gang Hua is with Dolby Laboratories, USA (e-mail: ganghua@gmail.com).}
}


\maketitle

\begin{abstract}
We introduce pluralistic salient object detection (PSOD), a novel task aimed at generating multiple plausible salient segmentation results for a given input image. Unlike conventional SOD methods that produce a single segmentation mask for salient objects, this new setting recognizes the inherent complexity of real-world images, comprising multiple objects, and the ambiguity in defining salient objects due to different user intentions. To study this task, we present two new SOD datasets ``\dataset'' and ``\datasetq'', along with newly designed evaluation metrics. \dataset~builds upon the DUTS dataset but enriches the ground-truth mask annotations from three aspects which 1) improves the mask quality especially for boundary and fine-grained structures; 2) alleviates the annotation inconsistency issue; and 3) provides multiple ground-truth masks for images with saliency ambiguity. \datasetq~consists of approximately 100K image-mask pairs with human-annotated preference scores, enabling the learning of real human preferences in measuring mask quality. Building upon these two datasets, we propose a simple yet effective pluralistic SOD baseline based on a Mixture-of-Experts (MOE) design. Equipped with two prediction heads, it simultaneously predicts multiple masks using different query prompts and predicts human preference scores for each mask candidate. Extensive experiments and analyses underscore the significance of our proposed datasets and affirm the effectiveness of our PSOD framework. 
\end{abstract}


\section{Introduction}
\begin{figure}[t]
   \centering
   \includegraphics[width=1\linewidth]{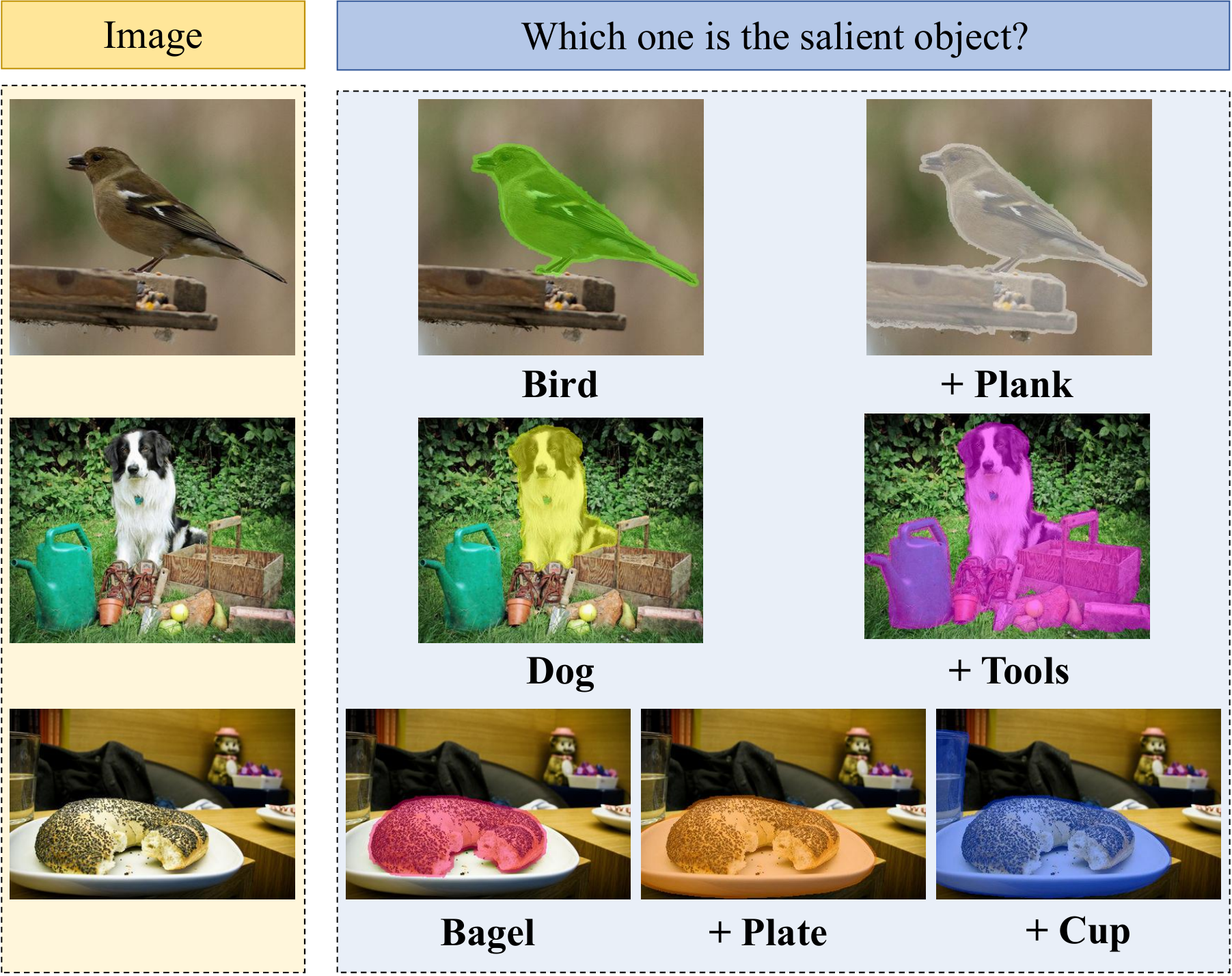}
   \caption{Three representative examples that illustrate the inherent ambiguity in defining salient objects. Salient object detection is an inherently ambiguous task. Therefore, one image with vague background or with more than two objects can have more than one salient region, which will result in more than one saliency maps.}
   \label{fig:motivation}
\end{figure}

\IEEEPARstart{S}{alient} object detection (SOD) is a classical vision task that seeks to automatically segment salient objects within a given input image. However, due to the inherent complexity of real-world images and varying user intentions, ambiguities often arise in defining salient objects. For instance, as shown in Fig.~\ref{fig:motivation}, when confronted with two or three objects in an image, segmentation becomes ambiguous. Objects on a dining table, such as food, plates and cups, may be perceived differently based on user intentions. For example, an individual desiring a drink may focus on the cup, while someone hungry may prioritize the food or even include its utensil. Conversely, an individual simultaneously thirsty and hungry may consider everything on the table as salient. This variability aligns with diverse user preferences, particularly in downstream SOD applications. 

\begin{figure}[t]
   \centering
   \includegraphics[width=1\linewidth]{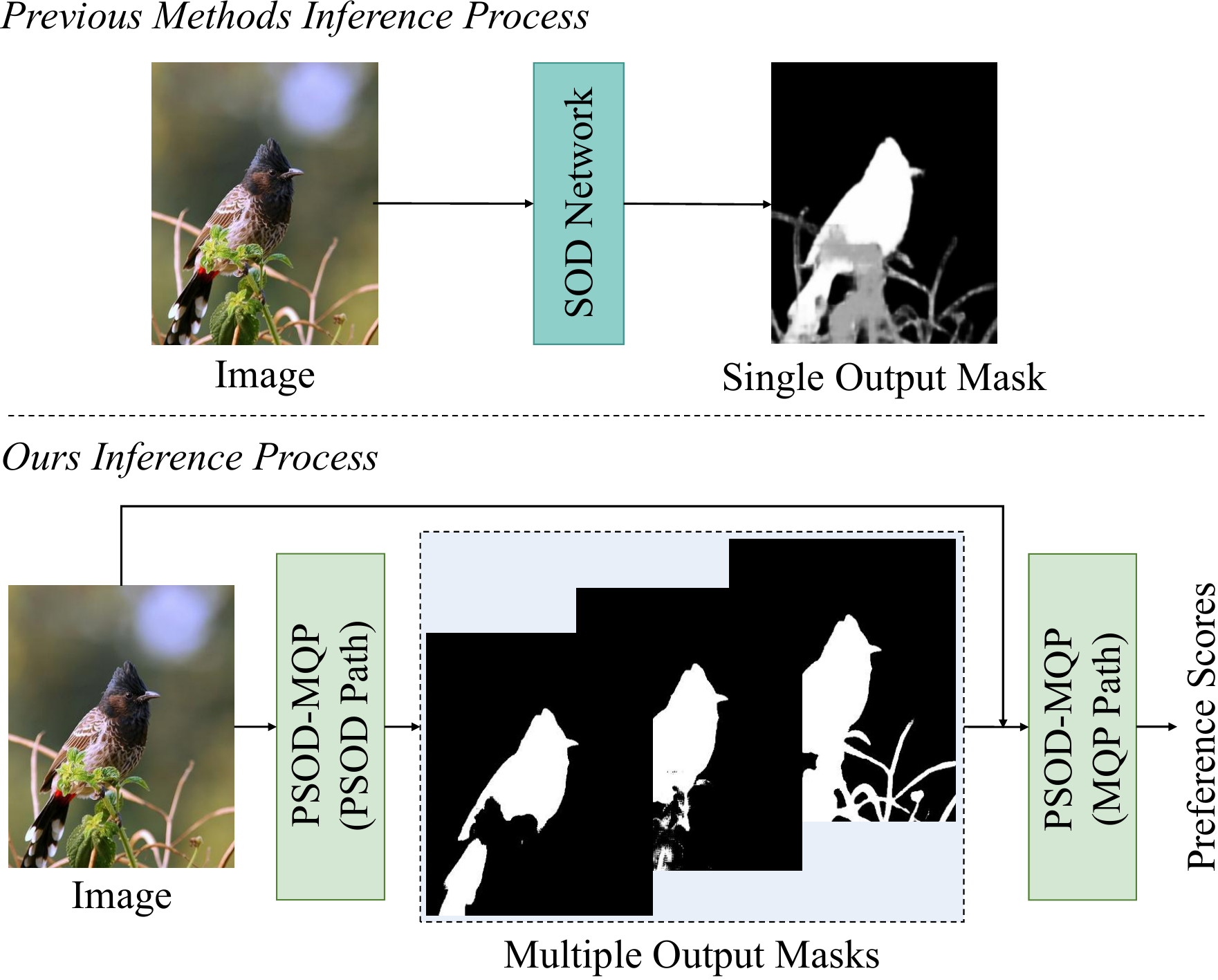}
   \caption{The inference process comparison between our method and previous SOD methods. When encountered images with ambiguity like the displayed image with blurred background, previous methods may predict grey in ambiguous regions, while our method outputs multiple masks and also grades the quality of masks during the inference stage.}
   \label{fig:inference}
\end{figure}


  

However, the inherent ambiguity in salient object detection (SOD) is overlooked by most existing datasets and methods, which treat SOD as a task with a singular solution. Under this task definition, existing datasets exhibit ambiguity issue when the datasets are labelled by multiple annotators with diverse intentions. Using the most widely utilized DUTS dataset~\cite{2017Learning} as an example, each image is associated with only a single ground-truth mask, despite a notable proportion of images featuring inherent ambiguity. As shown in Fig.~\ref{fig:visual_error}, the annotation inconsistency issue exists across many images. Such annotation inconsistency will also inevitably affect the performance of existing SOD methods which are designed to predict a single mask, as it introduces an ambiguous supervision signal. One typical illustrative example is shown in the top part of Fig.~\ref{fig:inference}. Here we want to emphasize that the inherent ambiguity of SOD is the root cause of the aforementioned annotation inconsistency issue and subsequent learning ambiguity. Only adopting an excessively strict and consistent annotation policy during dataset construction cannot fundamentally address the ambiguity issue arising from diverse user intentions. 

In this paper, we recognize the inherent ambiguity issue in SOD and propose pluralistic salient object detection (PSOD), shifting the task from single mask prediction to enabling generation of multiple salient mask candidates. To facilitate this, we introduce two new SOD datasets, ``\dataset'' and ``\datasetq''. \dataset~utilizes the same images as DUTS~\cite{2017Learning} but providing new enhanced annotations. Specifically, compared to the original DUTS annotations, our \dataset~improves in three key aspects: 1) Enhancing mask quality, particularly for boundary and fine-grained structures (\eg hair and bicycle wheels); 2) Addressing annotation ambiguity by providing multiple ground-truth masks for relevant images instead of a single one; 3) Alleviating annotation inconsistency issue existing in DUTS by adopting a more consistent annotation policy and permitting the annotation of multiple masks.

For \datasetq, it comprises about 100K image-mask pairs with human-annotated preference scores. Notably, the mask quality in this dataset exhibits a diverse distribution, encompassing both low-quality and high-quality masks. This dataset enables us to train a mask quality predictor, capable of predicting human preference without requiring knowledge of the ground-truth mask. It is very useful in some real user scenarios and automatically select high-quality ones. Moreover, considering existing segmentation metrics like mIoU may not always align with human preference, we find the learned mask quality predictor can potentially serve as an additional quality evaluator for existing SOD methods to show the alignment with real human feedback as well.

Building up the above two datasets, we further propose a simple yet effective end-to-end training baseline for pluralistic salient object detection, which can accomplish two tasks within one model, \ie outputting multiple mask candidates and predicting human preference scores for each mask candidate. Because these two tasks have different input/output formats and need to learn different representations, we find adopting two different input embedding tails and prediction heads while sharing the same backbone does not work well. Therefore, we introduce the mixture-of-experts (MOE) design and use different expert modules for each task in the shared backbone, which can well address the interference from each other. As shown at the bottom of Fig.~\ref{fig:inference}, our inference process involves generating multiple mask outputs initially. These masks pass through our MQP module in a batch, resulting in a preference score for each. The score can be leveraged as a guidance to decide whether the corresponding mask should be kept. Through comprehensive experimental results and analysis, we showcase the significant value of our newly introduced datasets and the efficacy of our PSOD baseline. Our key contributions can be summarized as follows:

\begin{itemize}
\item We introduce a novel task ``PSOD'', the first to address the inherent ambiguity in salient object detection, offering a fresh perspective and a redefined objective.
\item We present two large-scale datasets, \dataset~and \datasetq, curated for PSOD. These two datasets will be made publicly available, fostering further research along this direction. 
\item We present one simple yet effective end-to-end baseline for PSOD, capable of predicting not only multiple output mask candidates but also human preference scores. To our knowledge, this is also the first method that evaluates SOD quality in alignment with real human feedback.
\end{itemize}

\section{Related Work}
Over recent years, deep learning based methods \cite{wang2018salient, qin2019basnet, ma2023boosting, feng2022local, mukherjee2019dsal, jiang2019image} have achieved remarkable success in salient object detection. A crucial avenue for performance enhancement involves the incorporation of diverse network structures, such as fully convolutional networks \cite{wang2018salient, qin2019basnet, yang2019dilated, zhu2019aggregating, ma2023boosting,10102831}, Complex-valued Networks~\cite{jiang2019image}, and Transformers~\cite{liu2021visual, 10287608}. Additional directions for improvement encompass edge enhancement~\cite{zhao2019egnet, yang2022biconnet,liu2020dynamic}, feature extraction, and refinement~\cite{qin2020u2, jiang2019image, zhao2020suppress, zhang2018progressive, feng2019attentive}, lightweight model design~\cite{li2023rethinking, liu2021samnet}, as well as novel supervision strategies and loss functions~\cite{xu2021locate,li2021salient}. 

Along with the methodological development, a lot of SOD datasets have been introduced as well, including ECSSD~\cite{7182346}, PASCAL-S~\cite{6909437}, DUT-OMRON~\cite{6619251}, HKU-IS~\cite{li2015visual} and SOD~\cite{movahedi2010design}. Early datasets like SOD~\cite{movahedi2010design} and PASCAL-S~\cite{6909437} often have relatively small scales, covering only 300 and 850 images, respectively. ECSSD~\cite{7182346} offers 1,000 images, notable for their semantically rich yet intricate structures, accompanied by complex backgrounds. Recently, the most widely used dataset is DUTS~\cite{2017Learning}, boasting a larger scale with 10,553 training images and 5,019 test images sourced from ImageNet. Serving as the lifeblood and testbed, the evolution of these datasets has made a substantial contribution to the development of SOD techniques. Compared to our method, all above datasets and methodologies predominantly framed SOD as a single-mask prediction problem, overlooking its inherent ambiguity and diversity. In contrast, this paper introduces a distinct task definition along with two new datasets that permits the generation of multiple masks and prediction of human preference. 

Another SOD setting closely related to our proposed PSOD is salient object subitizing \cite{islam2018revisiting,zhang2015salient,he2017delving}, which predicts the number of salient object instances and potentially the saliency rank of each instance. The final saliency mask is then generated by merging one or multiple predicted object instance masks. However, this setting necessitates more costly instance-level object annotation and a more intricate network design, as it requires the model to differentiate between object instances. Contrastingly, the SOD task itself does not require such instance discrimination. Additionally, as segmenting high-quality object instance boundary is challenging, stitching multiple instance masks in this setting will result in unwanted hole/seam in the final mask. Furthermore, this setting struggles to handle salient regions belonging to stuff categories (\eg a tree in front of a blurred background), which are difficult to define as instances. Compared to salient object subitizing, our PSOD setting is notably simpler and does not need unnecessary instance differentiation, thus naturally capable of handling salient regions without clear instance definition.  

\section{Methods}
\label{sec:meth}

\begin{figure*}[t]
\begin{center}
\includegraphics[width=1.0\linewidth]{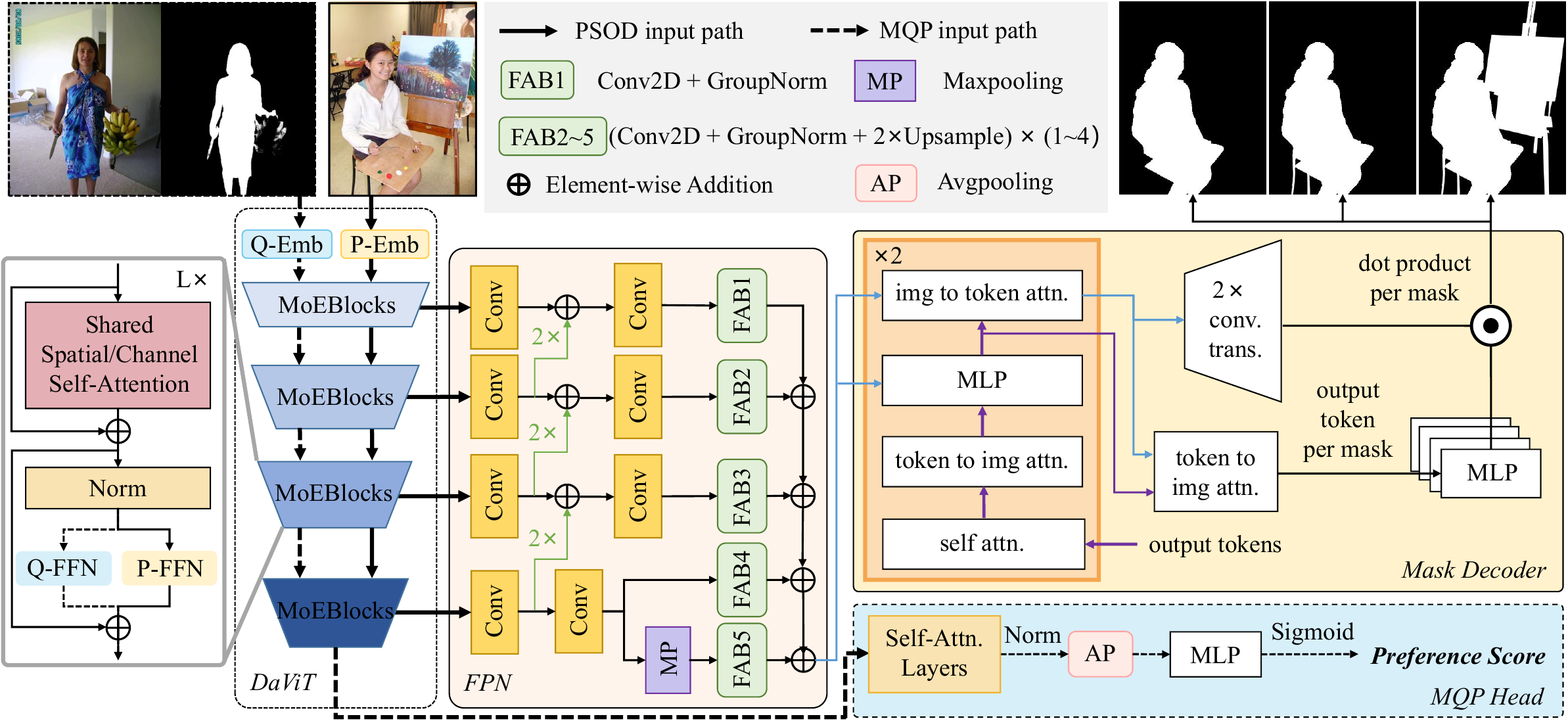}
\end{center}
   \caption{The overall architecture of our method, consisting of three parts: backbone encoder, FPN-based neck and prompt-driven mask decoder. To enable the model to handle two different tasks, we employ an MoE (Mixture of Experts) mechanism in the encoder.}
\label{fig:network}
\end{figure*}

Recognizing the inherent ambiguity of SOD, we design a simple yet effective baseline network ``PSOD-MQP'' for our newly proposed task setting. The overall framework is shown in Fig.~\ref{fig:network}. In order to predict human-preferred pluralistic saliency masks, it contains two sub-modules, \ie PSOD and mask quality predictor (MQP). PSOD is used to predict multiple masks to cover different potential saliency maps, while MQP is trying to predict the human preference score for each saliency mask by feeding the image-mask pair without knowing the groundtruth mask. Considering PSOD and MQP have different inputs (\ie image \textit{vs} image-mask pair) and require learning different characteristics, we find naively combining these two tasks within one model by using independent prediction heads and beginning embedding layers while sharing the same backbone does not work well. To address the learning interference, we incorporate Mixture-of-Expert (MoE) design into our transformer-based backbone as MoEBlocks. This is achieved by substituting the feed forward network (FFN) layer with an expert-switch layer, where two task-specific FFNs are trained for the two tasks respectively. In this paper, we simply adopt the DaViT\cite{ding2022davit} as the backbone by default, and other more advanced backbones \cite{liu2021swin,dong2022cswin} can also work well. And we replace the FFN layer in the original channel attention block and spatial attention block with a switch which applies task-specific FFN for PSOD and MQP. In Fig.~\ref{fig:network}, we denote PSOD expert and MQP expert as ``P-FFN'' and ``Q-FFN'', which are activated respectively when training the corresponding task. Besides the expert design, we also use different beginning embedding layers and prediction heads for the two tasks, which will be further elaborated below.

\subsection{PSOD Sub-Module}
Since current SOD methods predominantly focus on predicting a single saliency mask, they lack direct applicability for pluralistic salient object detection, which requires identifying multiple potential salient objects within an image. In this paper, inspired by the latest work SAM \cite{kirillov2023segment}, we design a prompt-driven mask decoder with multiple learnable output tokens as the PSOD head. 

In detail, given the input image, we first use one embedding layer to convert it into visual tokens. Then such tokens will be fed into the backbone to get the multi-scale features, where each expert-switch layer in the backbone MoEBlocks will select the P-FFN to process the incoming features. To aggregate the multi-scale features, we utilize a modified version of the FPN proposed in PanopticFPN \cite{kirillov2019panoptic} for segmentation tasks, by excluding the last upsampling layer. The output feature of the FPN is 1/4 of the input resolution and will be fed into the PSOD head to predict multiple mask candidates by being modulated by different learnable tokens via mutual cross-attention. It should be noted that we omit the IoU prediction branch in the SAM decoder \cite{kirillov2023segment} and use our proposed Mask Quality Predictor module to select the high-quality masks.

\subsection{MQP Sub-Module}
To predict the human preference score, we take an image and a mask candidate as the input and concatenate them along the channel dimension, and then use another embedding layer to convert them into visual tokens. Similarly, these visual tokens will be fed into the shared backbone while selecting the Q-FFN in each MoEBlocks. The output feature will be finally fed into the MQP prediction head to predict the preference score. In this paper, we simply design the MQP prediction head as a stack of self-attention layers, one global average pooling layer and one MLP layer followed by sigmoid. In real-world applications, users can leverage the MQP header to meet their specific requirements. For instance, if they prefer more diverse results, they can set a low score threshold, allowing the MQP to output more results. Conversely, if users prioritize mask quality, they can set a higher threshold, below which the corresponding masks are removed. In summary, the key benefits of MQP is that it offers a flexible solution that allows users to strike a balance between the quantity and quality of the results. This flexibility sets our method apart from other fixed mask approaches, emphasizing the benefits of integrating MQP into the whole PSOD framework.

\subsection{Losses}
To train our proposed model, we jointly train these two tasks and adopt different loss objectives accordingly. For PSOD task, we utilize the Cross-Entropy (CE) loss rather than the focal loss \cite{lin2017focal} used in SAM, since in our experiments we observe that focal loss is less stable. The possible reason is that, due to the existence of soft values in the ground-truth mask, the theoretical optimal point of focal loss is no longer at $p=t$, where $p,t$ is the prediction and ground-truth mask value, respectively. The Dice loss~\cite{milletari2016v} is kept and combined with the CE loss as follows:

\begin{equation}
\begin{gathered}
\mathcal{L}_{mask}= \lambda \mathcal{L}_{ce} + \mathcal{L}_{dice},
\end{gathered}
\end{equation}
where the value of $\lambda$ is assigned as 2.5 in our implementation. In addition, since multiple masks are generated, similar to SAM, we only backpropagate the mask that has the minimum loss to the ground-truth.

For MQP task, we treat it as a regression task, as our experiments empirically show improved performance and training stability compared to formulating it as a classification problem. One more benefit is the model can naturally predict a continuous preference score. By default, we simply use the Mean Squared Error (MSE) loss as our objective function, aiming to guide the model towards predicting preference scores that closely align with human-annotated scores.

\section{Dataset}

To support our PSOD setting, we further propose two new datasets, \ie \dataset~and \datasetq. \dataset~reuses the images from DUTS~\cite{2017Learning} dataset, but provides multiple possible saliency masks for images containing ambiguity in defining salient objects, rather than just one mask like DUTS. And \datasetq~provides massive image-mask pairs with human annotated preference scores under a diverse quality distribution. Before introducing our newly proposed datasets, we first analyze some annotations issues within DUTS.

\subsection{Annotation Issues within DUTS}
\begin{figure*}[t]
   \centering
   \includegraphics[width=1\linewidth]{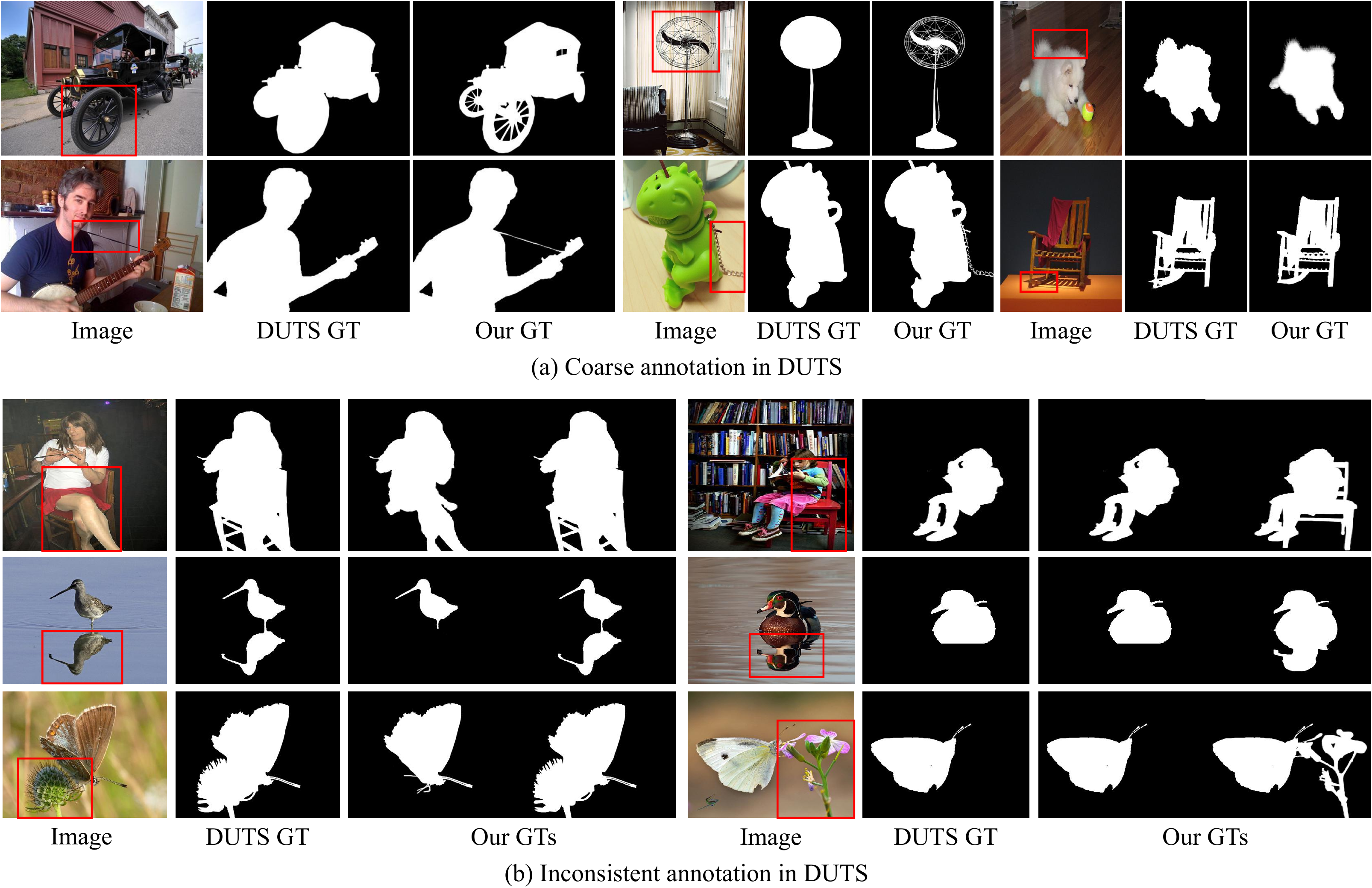}
   \caption{The visualization of coarse annotation issue (a) and inconsistent annotation issue (b) in DUTS and the comparison to our newly labeled \dataset~dataseet. We use red boxes to denote the coarsely/inconsistently annotated regions.}
   \label{fig:visual_error}
\end{figure*}

By examining the annotations within DUTS, we identify another two main issues: 1) coarse annotation quality; 2) annotation inconsistency. More specifically, the coarse annotation issue manifests in two aspects. First, some salient objects with thin structures are not annotated accurately. For example, in the first row of Fig.~\ref{fig:visual_error} (a), the wheel spokes and fan blades are not well segmented in the ground-truth mask, and the dog's hair is also delineated coarsely. Second, the issue of missing parts or introducing unrelated parts is widespread in DUTS. As shown in the second row of Fig.~\ref{fig:visual_error} (a), the guitar strap, which should be integral to the guitar, is missing. We observe that such coarse annotation issue occurs in both the training and test sets. Consequently, models trained on this dataset often fail in segmenting salient objects with thin structures, and this shortcoming is not evident from the final test performance.

Besides coarse annotations, DUTS also contains a considerable portion of images with inconsistent annotations. This inconsistency may stem from multiple human annotators participating in the annotation process, each with varying intentions or interpretations of saliency, especially in ambiguous cases. This actually further echoes the inherent ambiguity within SOD. As illustrated in Fig.~\ref{fig:visual_error} (b), the reflections of animals on water are perceived as salient objects, capturing attention with their mirrored beauty. However, in other instances, only the animals themselves are considered as the salient object. Due to the existence of such conflicted data annotation, the model is more likely to produce predictions with artifacts that exhibit low visual quality. Similar to the coarse annotation issue, inconsistent ground-truth in the test set can result in inaccurate test performance. To overcome these issues in the DUTS dataset, we conduct a comprehensive reannotation, providing finer masks and multiple ground-truth masks for images with ambiguity, which significantly alleviates the annotation inconsistency issue. We will discuss this further in the following sections.

\subsection{DUTS-MM Dataset}
\dataset~relabel images in the DUTS dataset that exhibits the above two issues and providing \textbf{m}ultiple \textbf{m}asks for images with inherent ambiguity. During the labeling process, we instruct annotators to provide masks as detailed as possible, especially around the boundary areas and thin structures of salient objects, to address the coarse annotation issue in DUTS. As for the inconsistent annotation, we observe that the primary cause is the ambiguous definition of a salient object. For instance, as shown in the left of the third row in Fig.~\ref{fig:visual_error} (b), both the individual and chair are collectively perceived as salient objects, yet in the right sample the salience is attributed exclusively to the person. Therefore, we ask annotators to provide multiple ground-truth masks for such images to encompass all possible combinations of salient objects. For images without ambiguity, we still annotate only one ground-truth masks. In Fig.~\ref{fig:visual_error} (b), we show some typical cases that contain ambiguity, including interaction between human and objects, reflection, and multiple objects in front of sky or blurred backgrounds.  

\begin{figure}[t]
    \centering
    \includegraphics[width=1\linewidth]{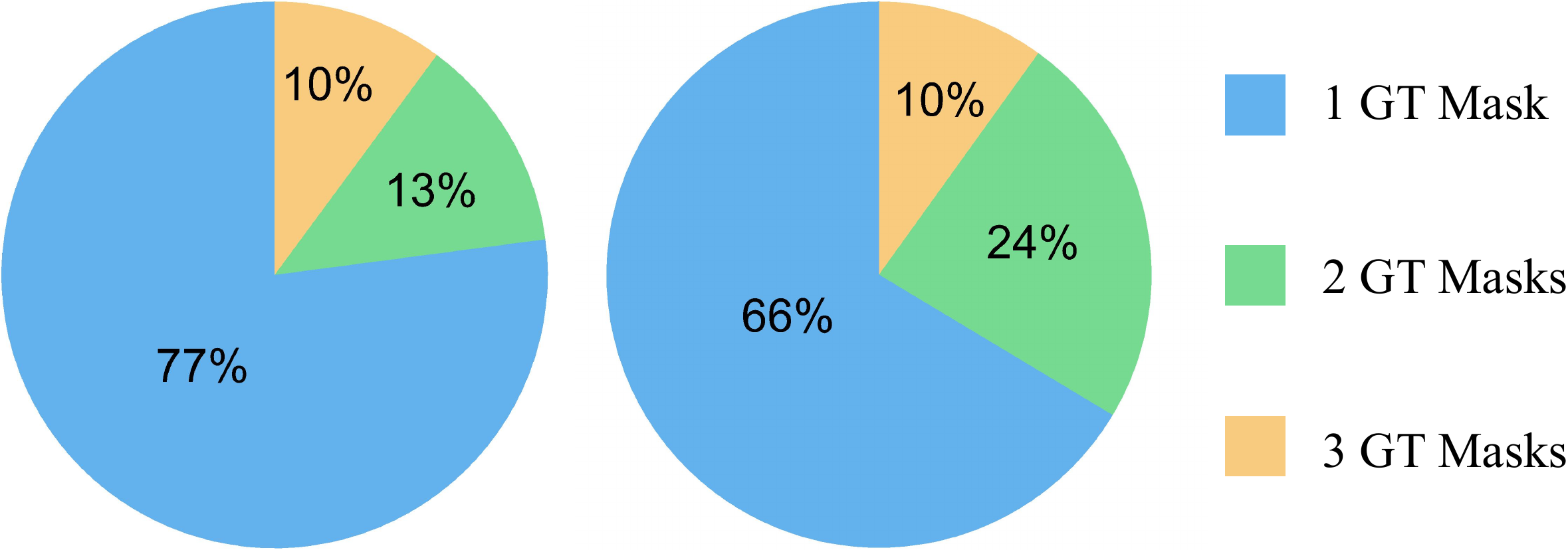}
    \caption{Distribution of the number of ground-truth (GT) mask each image has in the \dataset~training (top) and test (bottom) splits respectively.}
    \label{fig:mask_dis}
\end{figure}

\begin{figure}[t]
    \centering
    \includegraphics[width=1\linewidth]{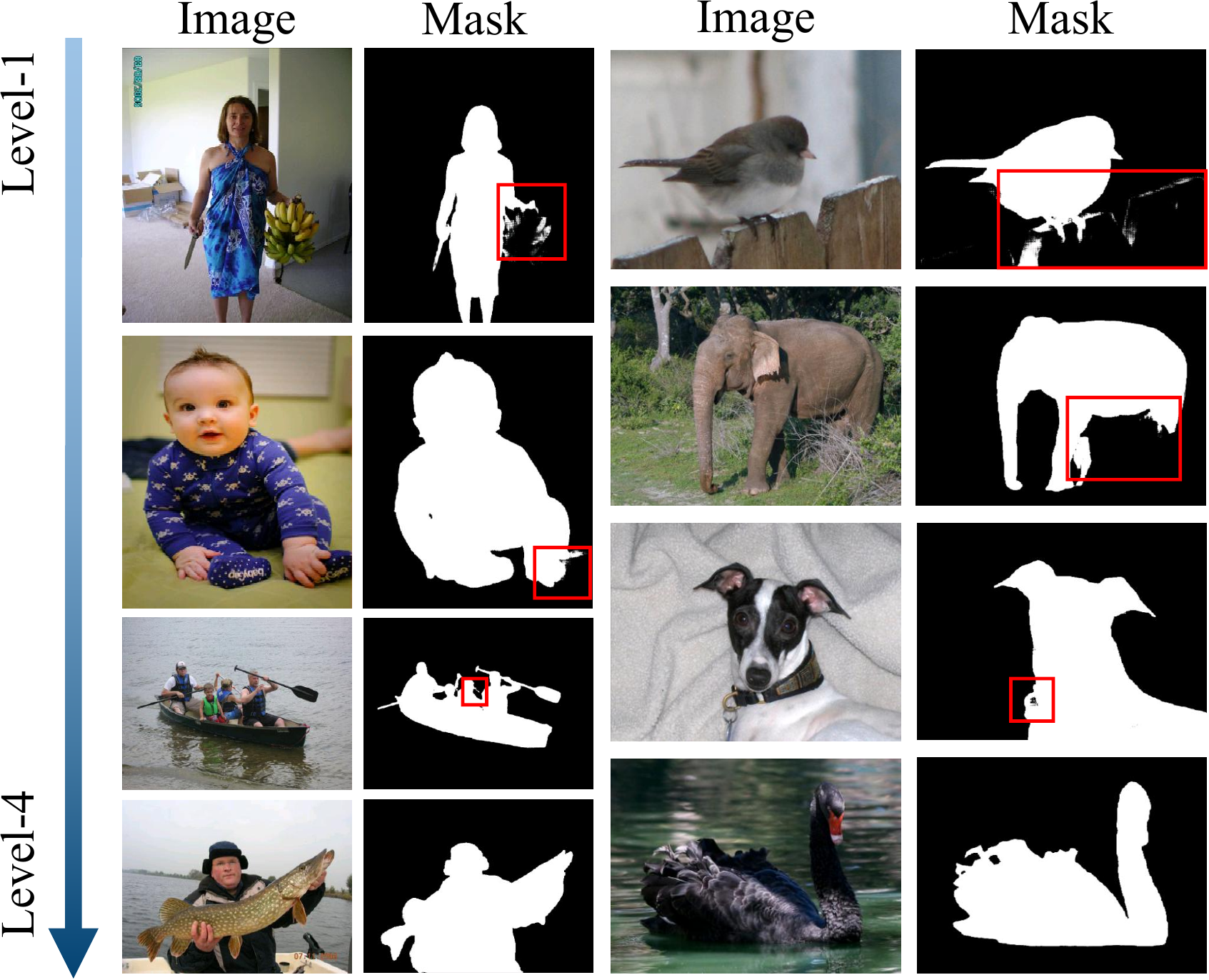}
    \caption{Examples of the \datasetq~Dataset. The image mask pairs are displayed in an ascending order of quality, starting from level 1 (lowest quality) at the top and progressing to level 4 (highest quality) at the bottom. We use the red box to segment low quality regions in each mask.}
    \label{fig:qualimask_visual}
 
\end{figure}

Detailed statistics of the number of ground-truth masks per image in our \dataset~dataset are illustrated in Fig.~\ref{fig:mask_dis}. Notably, approximately 23\% and 34\% of images contain more than one ground-truth mask in the training and test sets, respectively. Among these, a substantial proportion of images have three ground-truth masks. It is worth noting that the images in the DUTS dataset are meticulously selected to include salient objects. Through empirical observation, we find that three ground-truth masks suffice to cover all possible saliency masks for each of these images.

\subsection{DUTS-MQ Dataset}
Trained with the proposed \dataset~Dataset, our proposed PSOD task should be able to return multiple predicted masks per image. However, some of them may be not always with high quality, it would be super useful if we can automatically select the high-quality ones for users in real products. Since the ground-truth mask is unknown during testing, designing a quality predictor that can assess the mask quality without ground-truth reference become necessary.
To this end, we propose \datasetq~to support the training of such a \textbf{m}ask \textbf{q}uality \textbf{p}redictor. Each sample in \datasetq~is a triplet $\{\mathcal{I}, \mathcal{P}, \mathcal{Q}\}$, consisting of an input image $\mathcal{I}$, a predicted mask $\mathcal{P}$, and a human-annotated preference score $\mathcal{Q}$.  

In details, to build \datasetq, we carefully select images from the ImageNet dataset to maintain the similar distribution as DUTS, while avoiding identical images (to prevent the generation of perfect masks by models trained on DUTS). The image selection process involves computing the similarity between images in ImageNet and those in DUTS-TR based on features from the CLIP~\cite{radford2021learning} image encoder. We retain the images with similarities ranked second and third for each reference input from DUTS-TR, and then further filter them by manually excluding those without clear salient objects. 
For each selected image, we use two state-of-the-art SOD models (\ie TE3 and TE5 proposed in TRACER~\cite{lee2022tracer}) and our PSOD model (to be introduced in the following section) trained on \dataset~to generate different masks. Finally, we obtain 19,703 images, with each image containing 5 distinct masks. 

\begin{table}[t]
 \centering
 \caption{Distribution of the mask quality level (1-4) in the training and validation set of \datasetq~Dataset.}
\small
\begin{tabularx}{0.48\textwidth}{c|C|C|C}
\toprule
 Quality Level & Training & Validation & Total \\
\midrule
Level-1 & 31,489 & 1,874 & 33,363 \\
Level-2 & 30,129 & 1,801 & 31,930 \\
Level-3 & 15,963 & 965 & 16,928\\
Level-4 & 15,370 & 924 & 16,294 \\
\midrule
Total & 92,951 & 5,564 & 98,515 \\
\bottomrule
\end{tabularx}
\label{tab:sta}
\end{table}

In annotating mask quality, we instruct annotators to rely on their visual preferences rather than using mIoU as the criterion. This aims to simulate user feedback in real-world application scenarios. For each image-mask pair, annotators are asked to assign one of four distinct levels, ranging from $1$ to $4$, where $1$ and $4$ represents the least satisfactory and the ideal mask without any errors, respectively. Among the annotated image-mask pairs, we uniformly sample $5,564$ pairs as the validation set for determining training errors, while the remaining pairs form the training set. Some typical annotation examples are shown in Fig.~\ref{fig:qualimask_visual}. It can be seen that, some masks assigned level 1, despite having high mIoU values, leave a poor impression to humans due to annoying artifacts. But some masks missing a large part (elephant in the second row) are even more preferable. Need to note that both predicted masks with artifacts are typical examples of salient ambiguity cases we discussed earlier, likely caused by inconsistent annotations. This observation highlights the necessity of addressing the inherent ambiguity and annotation inconsistency in SOD. The detailed statistics of \datasetq~are shown in Table~\ref{tab:sta}. It reveals that masks with quality levels of $1$ and $2$ account for more than $65\%$ of the entire dataset, while only $16\%$ of the samples (annotated to level $4$) are considered ``perfect'' by the human annotators, indicating a substantial room for improvement in current SOD models.

Need to note that, when annotating the above two datasets, we have two annotators to do cross-check, one annotating the mask/score and another reviewing the annotation result to guarantee the quality. If the reviewer is uncertain about some hard cases, the third senior supervisor will be involved and correct the annotation labels if needed.

\subsection{Evaluation Metrics}
Since multiple prediction masks and ground-truth masks exist in the PSOD setting, we can no longer rely on conventional one-one matching metrics such as mIoU, which are employed in previous SOD studies. In this paper, we redefine the average precision $AP$, average recall $AR$ and $F_1$ score to evaluate the model for the PSOD task. Specifically, given $K$ predicted masks ({\footnotesize $\hat{M}_{i_{1}},...,\hat{M}_{i_{K}}$}) and $J$ GT masks ({\footnotesize $M_{i_{1}},..., M_{i_{J}}$}) for the $i$th image, the precision is defined as:

\begin{equation}
Prec_i = \frac{1}{K}\sum_{k=1}^K\mathcal{M}(\hat{M}_{i_{k}},M_{i_{j}}^*),
\end{equation}
where {\footnotesize $\mathcal{M}(\cdot)$} denotes the average of mean F-measure~\cite{margolin2014evaluate} and S-measure~\cite{fan2017structure}. {\footnotesize $M_{i_{j}}^*=\underset{j \in {1,2,..,J}}{\arg \max }\mathcal{M} (\hat{M}_{i_{k}}, M_{i_{j}})$ }that represents the ground-truth best matches the predicted mask {\footnotesize $\hat{M}_{i_{k}}$}. Similarly, the recall is defined as:
\begin{equation}
Rec_i = \frac{1}{J}\sum_{j=1}^J\mathcal{M}(\hat{M}_{i_{k}}^*,M_{i_{j}}),
\end{equation}
where {\footnotesize $\hat{M}_{i_{k}}^*=\underset{k \in {1,2,..,K}}{\arg \max }\mathcal{M} (\hat{M}_{i_{k}}, M_{i_{j}})$} represents the closest predicted masks to the ground-truth. Finally, we calculate the average precision ($AP$) and average recall ($AR$) of $N$ images, and $F_1$ score is the harmonic mean of $AP$ and $AR$.

\section{Experiments}
\label{sec:exp}
\subsection{Implementation Details.} 
We employ the ImageNet \cite{ILSVRC15} pretrained DaViT-Tiny \cite{ding2022davit} as the backbone by default. For PSOD, the FPN neck and mask decoder are initialized randomly. We scale inputs to the resolution of ${512 \times 512}$ for both training and inference phases. The output channel of the FPN features is set to $256$ while the mask decoder is designed to generate $5$ masks per image. Aligning with PSOD sub-module, inputs (concatenation of an image and a predicted mask) of MQP are scaled to ${512 \times 512}$ as well. We utilize the Adam optimizer with beta1 and beta2 values of $0.5$ and $0.999$, respectively. The initial learning rate is set at $0.0001$. For PSOD, it follows a cosine learning rate scheduler decaying to $0$, whereas for MQP, it is decreased by a factor of $10$ if the validation set loss plateaus over five consecutive epochs. The model is trained for $50$ epochs with batch size 4/8 (PSOD/MQP) on $4$ RTX 2080Ti GPUs. For MQP training, we normalize the ground-truth scores by mapping the quality level \{1, 2, 3, 4\} to \{0, 0.33, 0.67, 1.0\} respectively. During the inference stage, we initially execute a forward pass to generate multiple saliency maps, followed by a second forward pass specifically dedicated to assessing the quality of each generated mask.

\begin{table}[t]
 \centering
 \caption{Comparison results with existing state-of-the-art SOD methods, which shows that our proposed PSOD can produce superior performance.}
\label{tab:sota}
 \setlength{\tabcolsep}{5.8mm}{
\begin{tabular}{c|c|c|c}
\toprule
Method & $AP$ $\uparrow$ & $AR$ $\uparrow$ & $F_1$ $\uparrow$ \\
\midrule
\multicolumn{4}{c}{State-of-the-Art SOD Methods} \\
\midrule
MINet~\cite{MINet-CVPR2020} & 0.853 & 0.826 & 0.839 \\
VST~\cite{liu2021visual} & 0.857 & 0.829 & 0.843 \\
TRACER~\cite{lee2022tracer} & 0.878 & 0.841 & 0.859  \\
EDN~\cite{wu2022edn} & 0.853 & 0.826 & 0.839 \\
SelfReformer~\cite{10287608} & 0.873 & 0.841 & 0.857\\
\midrule
\multicolumn{4}{c}{State-of-the-Art High-Resolution SOD Methods} \\
\midrule
PGNet \cite{xie2022pyramid} & 0.862 & 0.831 & 0.846 \\
InSPyReNet \cite{kim2022revisiting} & 0.889 & 0.857 & 0.873 \\
\midrule
\multicolumn{4}{c}{Open-sourced SOD via Subitizing Methods} \\
\midrule
RSDNet \cite{Islam_2018_CVPR} & 0.743 & 0.722 & 0.732 \\
\midrule
\multicolumn{4}{c}{Our Method} \\
\midrule
Our PSOD & \textbf{0.889} & \textbf{0.904} & \textbf{0.896}\\ 
 \bottomrule
\end{tabular}
}
\end{table}

\begin{table}[h!]
 \centering
\caption{Performance Comparison of Different Methods on Four Benchmark Datasets: DUT-OMRON, ECSSD, HKU-IS, and PASCAL-S.}
\begin{tabularx}{0.48\textwidth}{l|CCCCCCCC}
\toprule
		\multirow{2}{*}{Method} & \multicolumn{4}{c}{DUT-OMRON}& \multicolumn{4}{c}{ECSSD}  \\
		\cmidrule(lr){2-5} \cmidrule(lr){6-9}
		& $F_\beta^{max}$ & $F_\beta^{avg}$ & $S_m$ & MAE & $F_\beta^{max}$ & $F_\beta^{avg}$ & $S_m$ & MAE \\
		\midrule
        TRACER \cite{lee2022tracer} & .853 & .805 & .861 & .047 &  .960 & .938 & .937 & .024 \\
        VSCode \cite{luo2023vscode} & - & .830 & .869 & - & - & .957 & .945 & - \\
        Our PSOD & \textbf{.867} & \textbf{.849} & \textbf{.887} & \textbf{.031} & \textbf{.968} & \textbf{.958} & \textbf{.949} & \textbf{.017}\\
		\midrule
        \midrule
		\multirow{2}{*}{Method} & \multicolumn{4}{c}{HKU-IS} & \multicolumn{4}{c}{PASCAL-S} \\
		\cmidrule(lr){2-5} \cmidrule(lr){6-9}
		& $F_\beta^{max}$ & $F_\beta^{avg}$ & $S_m$ & MAE & $F_\beta^{max}$ & $F_\beta^{avg}$ & $S_m$ & MAE \\
		\midrule
        TRACER \cite{lee2022tracer}  &.952  & .923 & .933 & .020  & \textbf{.901}  & .865  & .880 &  .048 \\
        VSCode \cite{luo2023vscode} & -  & \textbf{.946} & .935 & -  & -  & .852  & .878 &  - \\
        Our PSOD & \textbf{.955} & \textbf{.943} & \textbf{.940} & \textbf{.017} & .895 & \textbf{.878} & \textbf{.884} & \textbf{.043} \\
		\bottomrule
\end{tabularx}
\label{tab:benchmark}
\end{table}

\subsection{Main Experiment Results}
\paragraph{\textbf{Quantitative Comparison}}
In this section, we evaluate our PSOD model on the proposed \dataset~dataset and compare our results with five state-of-the-art SOD methods, namely MINet~\cite{MINet-CVPR2020}, VST~\cite{liu2021visual}, TRACER~\cite{lee2022tracer}, EDN~\cite{wu2022edn}, and SelfReformer~\cite{10287608}. In addition, we also compare with two SOTA high-resolution SOD methods (PGNet~\cite{xie2022pyramid} and InSPyReNet~\cite{kim2022revisiting}) and the only one open-sourced\&runnable subitizing method RSDNet~\cite{Islam_2018_CVPR}. We keep the original experimental settings of these methods. All the methods are retrained and evaluated on the \dataset. Besides, we retrain two high-resolution SOD methods with the same resolution ($512 \times512$) and mask quality. There is no training mask annotation quality difference. The only difference is PSOD is supervised by multiple GTs at each iteration while others are learnt with one of GT masks (one GT mask will be sampled at each iteration, and multiple GT masks will be regarded as multiple training samples) due to their architecture constrain. The main results are displayed in Table~\ref{tab:sota}. The results of our model are derived by evaluating the masks generated and filtered by the PSOD module and MQP module respectively. For mask filtering, we establish a threshold for the mask quality predictor, below which the corresponding mask is deemed as `poor' and excluded from evaluation. By adjusting this threshold, we obtain and report the optimal $F_1$ score along with its corresponding $AP$ and $AR$ in Table \ref{tab:sota}. 

As observed, our PSOD network + MQP framework significantly surpasses existing methods, improving $AP$, $AR$, and $F_1$ by margins of $1.1\%$, $6.3\%$, and $3.7\%$ over the best-performing method, respectively. And for result comparing with HRSOD, it demonstrates PSOD still has a clear superiority in terms of $AR$ and $F_1$. Moreover, it shows our method clearly beats the performance of subitizing-based SOD method.

\begin{figure}[t]
    \centering
    \includegraphics[width=1\linewidth]{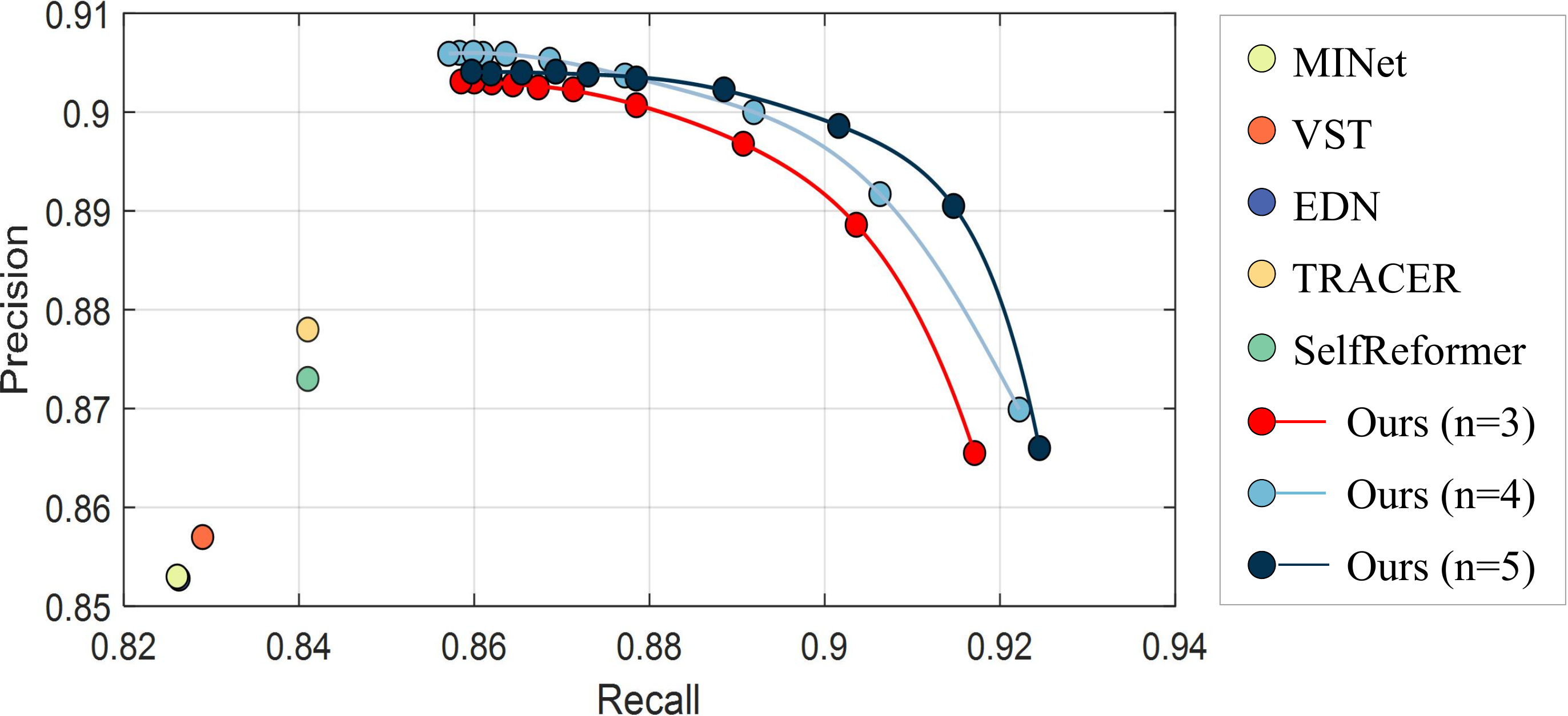}
    \caption{Precision-Recall Curve for SOD methods. The proposed method has the overall best performance. For our methods, we show three curves obtained by adapting the number of output masks ($n=3,4,5$) given by the PSOD sub-module.}
    \label{fig:pr}
\end{figure}

To more effectively show the superiority of our framework, we construct a precision-recall curve in Fig.~\ref{fig:pr}. The curves are generated by adjusting the threshold for the MQP from $0$ to $0.9$ in increments of $0.1$. Here only the curve obtained by the PSOD model with $5$ output masks ($n=5$ in Fig.~\ref{fig:pr}) is discussed. As observed, our curve lies to the right of all other state-of-the-art methods, indicating superior performance. Furthermore, the $AP$ range spans from $0.866$ to $0.904$, while the $AR$ range extends from $0.924$ to $0.860$. This demonstrates that by incorporating the MQP, our system can more effectively balance precision and recall to meet the requirements of various users, whereas other methods produce fixed masks. In other words, in real-world applications, users can adjust the threshold based on their preferences to obtain more tailored results.

To further verify the generalizability of our model across different datasets, we conducted a comprehensive comparison with the top-performing state-of-the-art methods, TRACER and VSCode-T~\cite{luo2023vscode}, using four additional benchmark datasets: DUT-OMRON, ECSSD, HKU-IS, and PASCAL-S. The results, as presented in Table~\ref{tab:benchmark}, demonstrate that our method outperforms the others on most of these benchmarks across prevalent evaluation metrics. This indicates the robustness and effectiveness of our approach in diverse scenarios.

\begin{figure}[t]
    \centering
    \includegraphics[width=1\linewidth]{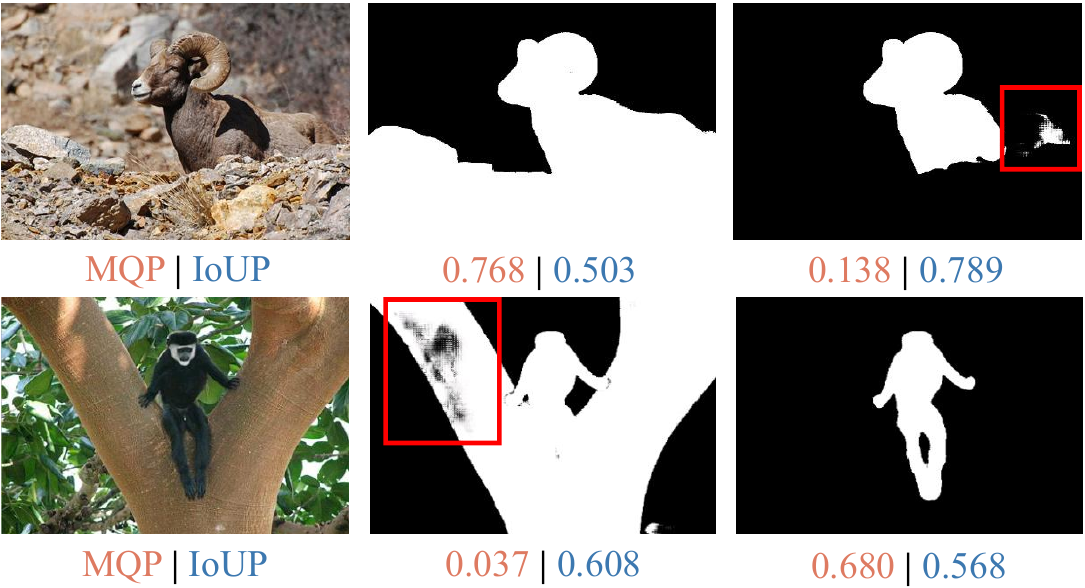}
    \caption{Examples of mask quality prediction between our MQP and IoU Predictor from SAM~\cite{kirillov2023segment}. The red box is used to mark the low-quality region in the mask. The higher the value, the better the quality of the mask.}
    \label{fig:visual_mqp_ioup}
\end{figure}

\begin{figure*}[t]
\begin{center}
\includegraphics[width=1\linewidth]{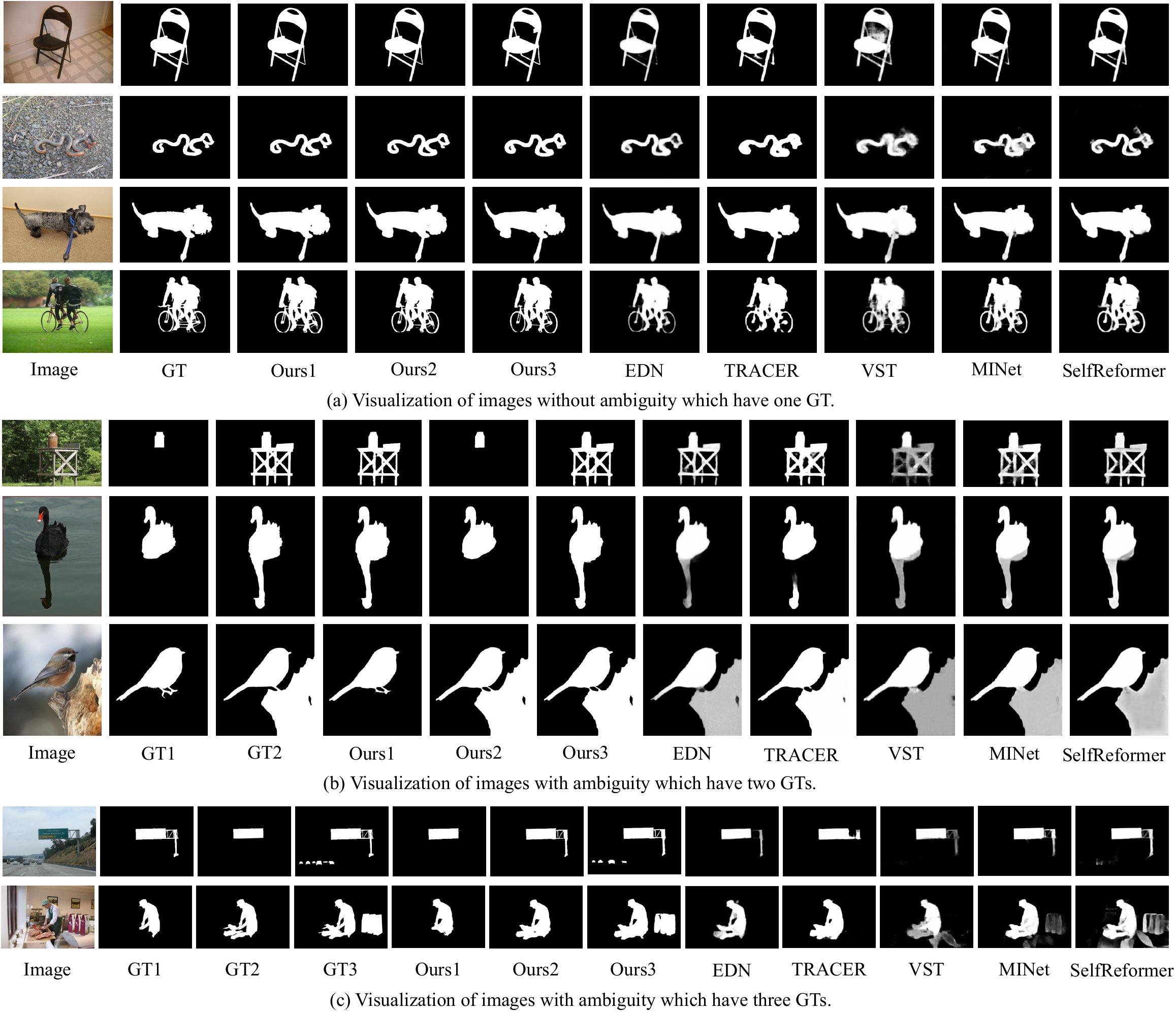}
\end{center}
   \caption{Visual comparison with state-of-the-art methods. Our method can achieve better and diverse visual results on both images without and with ambiguity.}
\label{fig:visualsota}
\end{figure*}

\begin{figure*}
\begin{center}
\includegraphics[width=1\linewidth]{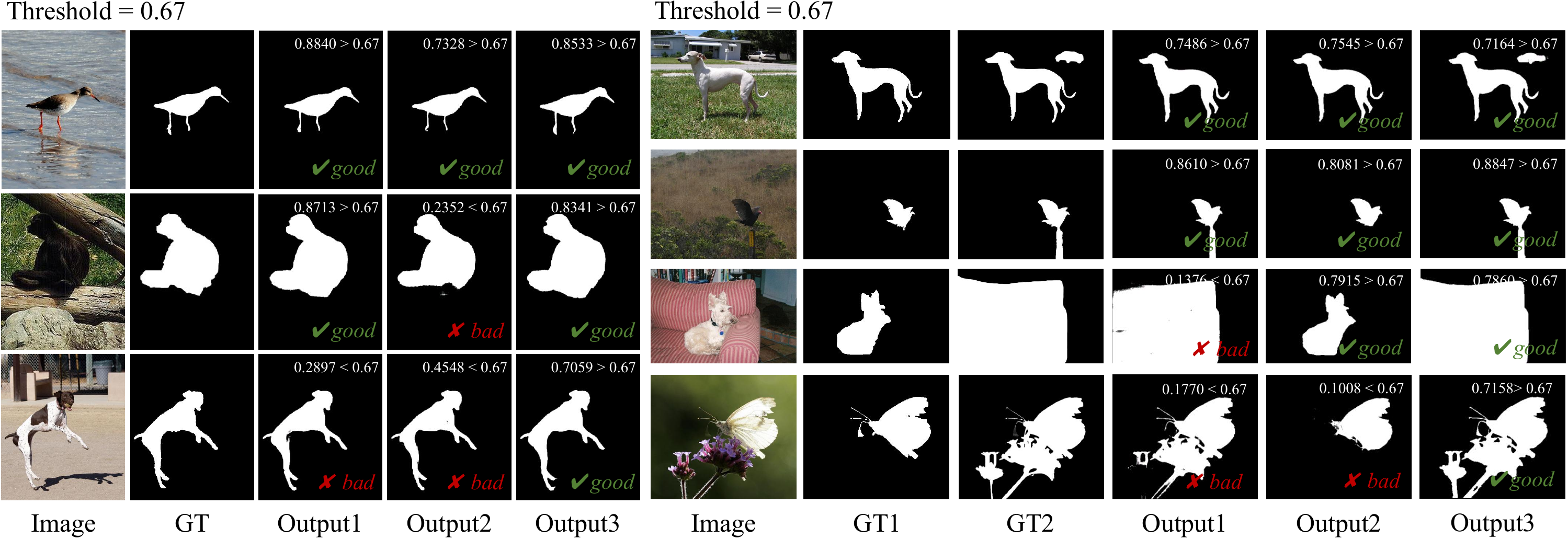}
\end{center}

   \caption{Visualization on the score generated from MQP corresponding to multiple prediction masks from PSOD. Here we show the case by setting quality threshold as 0.67 to filter out low-quality masks. The left and right part show the output score of images without and with ambiguity respectively.}
\label{fig:visualmqp}
\end{figure*}

\paragraph{\textbf{Qualitative Comparison}}
In addition to the quantitative results, Fig.~\ref{fig:visualsota} shows
the qualitative results of our PSOD model and 5 other methods over three representative images in \dataset~test Dataset. We display $3$ output masks selected by the MQP per image. We can see from the figure that our method is capable of generating different saliency maps. And for each map, our model well locate the salient regions, making the predicted saliency map closer to the ground-truth than other methods. We also visualize the output scores of mask quality predictor for multiple prediction mask of PSOD in Fig.~\ref{fig:visualmqp} (similar to Fig.~\ref{fig:visualsota}, only masks with top-3 MQP score are shown). By setting proper quality threshold, our mask quality predictor can help filter out low-quality masks for users automatically.

\subsection{Ablation Study}
\paragraph{\textbf{Impact of Model Size}}
We compare multiple backbone encoders with varying model sizes, ranging from DaViT-Tiny (28.3 Mb) to DaViT-Base (87.9 Mb). The results are presented in Table~\ref{tab:ab1}. It can be seen that increasing the model size results in only minor performance improvements, while introducing more computation cost. Therefore, we utilize DaViT-Tiny throughout the paper to maintain efficiency by default. If we further increase the dataset scale and complexity, larger models may help more. 

\begin{table}[t]
\centering
\caption{Ablation analysis for different sizes of backbone in PSOD network.}
\label{tab:ab1}
\begin{tabularx}{0.48\textwidth}{c|C|C|C|C}
\toprule
Backbone & \#Params & $AP$ $\uparrow$ &  $AR$ $\uparrow$ & $F_1$ $\uparrow$ \\
\midrule
DaViT-Tiny & 28.3M & 0.889 &  0.904 &  0.896  \\
DaViT-Small & 49.7M & 0.889 & 0.903 &  0.896 \\
DaViT-Base & 87.9M & \textbf{0.898} & \textbf{0.905} & \textbf{0.902} \\
\bottomrule
\end{tabularx}
\end{table}

\paragraph{\textbf{Performance gain from backbone?}}
In Table~\ref{tab:ab4}, we substitute the backbone networks in TRACER (EfficientNet), EDN (ResNet50), and SelfReformer (PVT) with our consistent backbone DaViT-Tiny, allowing for a direct comparison of performance across different methods using the same backbone. Despite this uniform modification, our PSOD method continues to exhibit superior performance, highlighting the robustness and effectiveness of our approach. Specifically, our method's $AP$ score is higher by 0.052, its $AR$ score is higher by 0.094, and its $F_1$ score is higher by 0.073. These results underscore the strength of our PSOD approach, as it consistently delivers higher accuracy, recall, and $F_1$ scores, even when the same backbone network is employed, demonstrating its superior performance across all evaluated metrics. 

\begin{table}[t]
\centering
\caption{Ablation analysis for different SOD methods using DaViT-Tiny as backbone network.}
\label{tab:ab4}
\begin{tabularx}{0.48\textwidth}{l|C|C|C}
\toprule
Method  & $AP$$\uparrow$ & $AR$$\uparrow$  & $F_1$ $\uparrow$\\
\midrule
EDN \cite{wu2022edn} & 0.837 & 0.810 & 0.823 \\
TRACER \cite{lee2022tracer} & 0.871 & 0.837 & 0.854 \\
SelfReformer \cite{10287608} & 0.867 & 0.831 & 0.849 \\
Our PSOD & \textbf{0.889} & \textbf{0.904} & \textbf{0.896}\\
\bottomrule
\end{tabularx}
\end{table}

\begin{table}[t]
\centering
\caption{Ablation analysis for using different number of output masks in the mask decoder.}
\label{tab:ab2}
\begin{tabularx}{0.48\textwidth}{c|C|C|C}
\toprule
Output (No.) & $AP$ $\uparrow$ &  $AR$ $\uparrow$ & $F_1$ $\uparrow$\\
\midrule
$n=3$ & 0.889 & 0.904 & 0.896\\
$n=4$ & \textbf{0.892} & 0.906 & 0.899 \\
$n=5$ & 0.891 & \textbf{0.915} & \textbf{0.902}\\
$n=6$ & 0.875 & 0.907 & 0.891 \\
\bottomrule
\end{tabularx}
\end{table}

\paragraph{\textbf{Number of output masks of PSOD}}
By using more output tokens as prompt, Our PSOD framework can output more mask candidates. Here we conduct an ablation study on the number of output masks generated by the proposed PSOD. In Table \ref{tab:ab2} and Fig. \ref{fig:pr}, we show the results with $3\sim 6$ output masks. We can observe that generating more masks will yield slightly better results when $n$ is changed from $3$ to $5$ and the performance drops at $n=6$. As adding an output only introduces a little computation cost increase related to the total cost, we opt for $5$ output masks for PSOD.

\paragraph{\textbf{MQP vs. SAM IoU Predictor}}
In this paper, we employ an additional mask quality predictor module to autonomously assess the quality of predicted masks. Notably, SAM \cite{kirillov2023segment} integrates an IoU predictor. To show the superiority of MQP over the IoU predictor in reflecting real human preference, we train a variant of PSOD with an additional IoU predictor branch. For quantitative comparison, we randomly select $500$ images from the \dataset~test set. For each image, we choose a pair of predicted masks with visually discernible quality differences from the output of PSOD. We then ask human annotators to assign a score of $1$ to the superior mask and $0$ to the inferior one, establishing a ground-truth (GT). The output from both MQP and IoU prediction scores is subsequently employed to rank each mask pair and assess consistency with human feedback. On average, MQP achieves an accuracy of $94.2\%$, surpassing SAM's IoU predictor by $12.4\%$ ($94.2\%$ vs. $81.8\%$). One visual comparison example is provided in Fig.~\ref{fig:visual_mqp_ioup}. It is evident that MQP is more sensitive to visual flaws in the masks, assigning significantly lower scores, whereas SAM's IoU predictor focuses more on IoU, which cannot accurately reflect the human preference.

\begin{figure}
    \centering
    \includegraphics[width=1\linewidth]{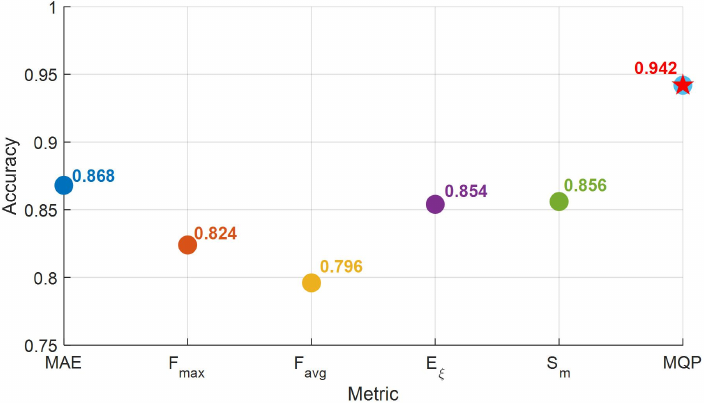}
    \caption{Ablation analysis for comparing MQP with different evaluation metrics in SOD to evaluate mask quality.}
    \label{fig:visual_ab5}
\end{figure}

\paragraph{\textbf{MQP vs. Current Metrics}}
To validate the efficacy of our MQP sub-module in assessing mask quality that aligns with human preference, we conduct a thorough comparison between preference scores from MQP and existing established SOD evaluation metrics. Specifically, we compare MQP against five widely used metrics, \textit{i.e.}, mean absolute error (MAE)~\cite{MINet-CVPR2020}, max F-measure ($F_{max}$)~\cite{MINet-CVPR2020}, mean F-measure ($F_{avg}$)~\cite{MINet-CVPR2020}, mean E-measure ($E_\xi$)~\cite{fan2018enhanced} and S-measure ($S_m$)~\cite{fan2017structure}. Apart from MAE where smaller is better, the other four metrics are better when larger. We conduct this experiment under the same setting when comparing MQP with SAM's IoU predictor. The alignment accuracy with human feedback is presented in Fig.~\ref{fig:visual_ab5}. It shows that our MQP sub-module clearly outperforms existing SOD evaluation metrics, demonstrating closer alignment with the real human feedback.


\paragraph{\textbf{Enhancing performance and human preference alignment using MQP}}
MQP enables automatic ranking not only for PSOD but also for existing SOD methods in real products. To show this capability, we use existing five existing SOTA SOD methods to generate the SOD masks for randomly chosen 500 images from \dataset, and then use MQP to choose the best SOD mask for each image without knowing the GT SOD mask. This automatic selection boosts the F1 evaluation metric to 0.87, which is better than the best-performing method (F1: 0.86). More importantly, the selected masks align with human preference much better (alignment acc: 77.5\%) than mask selection using existing evaluation metrics ($F_\beta^{avg}$) that require GT masks (alignment acc: 57.0\%), which is not available in real products.

\begin{table}[t]
\centering
\caption{Ablation analysis for fine-tuning on EfficientSAM to verify the effectiveness of our PSOD in segmenting salient objects.}
\label{tab:ab3}
\begin{tabularx}{0.48\textwidth}{c|C|C|C}
\toprule
Method  & $AP$ $\uparrow$ & $AR$ $\uparrow$ & $F_1$ $\uparrow$\\
\midrule
EfficientSAM \cite{xiong2023efficientsam} & 0.851 & 0.879 & 0.865 \\
PSOD (n=3) & \textbf{0.889} & \textbf{0.904} & \textbf{0.896}\\
\bottomrule
\end{tabularx}
\end{table}

\paragraph{\textbf{PSOD vs. SAM}} We conduct an experiment to fine-tune the EfficientSAM-S~\cite{xiong2023efficientsam} on our dataset, given the huge model size of SAM. In detail, we employ a full image bounding box to prompt the model. The results can be seen in Table~\ref{tab:ab3}. Interestingly, merely fine-tuning SAM does not yield superior performance. This might because SAM is not specifically designed and pre-trained for SOD task.

\section{Conclusion}
\label{sec:conc}
This paper pioneers the exploration of inherent ambiguity in Salient Object Detection (SOD) by introducing a novel perspective: redefining SOD as a pluralistic salient object detection (PSOD) task, generating multiple mask candidates per image. To facilitate the training of PSOD, we introduce two new large-scale datasets, \dataset~and \datasetq. Different from existing SOD datasets, which offer only one ground-truth mask per image, \dataset~is tailored to supply multiple saliency masks for images containing ambiguity in defining salient objects. Additionally, we enhance annotation quality and mitigate issues of annotation inconsistency within existing SOD dataset. Notably, \datasetq~is the first dataset designed to provide real human visual preference scores for evaluating mask quality, without the need for knowledge about the ground-truth mask. Building on these two new datasets, we propose a simple PSOD baseline capable of predicting multiple masks in SOD and human preference scores for each mask, which can be used to automatically estimate the mask quality aligning with real human feedback. We believe our paper will provide a fresh perspective to the research community and inspire more great works along this direction.


%
%

\bibliographystyle{splncs04}
\bibliography{main}

\end{document}